\documentclass{article}

\usepackage{tikz}
\usetikzlibrary{positioning, arrows.meta, fit, backgrounds}
\usepackage{graphicx}
\usepackage{amsfonts}
\usepackage{amsmath}
\usepackage[hidelinks,colorlinks=true,linkcolor=blue,citecolor=blue]{hyperref}
\usepackage{subcaption}
\usepackage{xcolor}
\usepackage[mathscr]{euscript}
\usepackage[a4paper, left=1in, right=1in, top=1in, bottom=1in]{geometry}
\usepackage{authblk}
\usepackage{appendix} 
\usepackage{booktabs} 

\title{{FluidFlow}: a flow-matching generative model for fluid dynamics surrogates on unstructured meshes}

\author[1]{David Ramos}
\author[2]{Lucas Lacasa}
\author[1]{Fermín Gutiérrez}
\author[1,3]{Eusebio Valero}
\author[1,3]{Gonzalo Rubio}

\affil[1]{ETSIAE-UPM-School of Aeronautics, Universidad Politécnica de Madrid, Plaza Cardenal Cisneros 3, E-28040 Madrid, Spain}
\affil[2]{Institute for Cross-Disciplinary Physics and Complex Systems (IFISC, CSIC-UIB), 07122 Palma de Mallorca (Spain)}
\affil[3]{Center for Computational Simulation, Universidad Politécnica de Madrid, Campus de Montegancedo, Boadilla del Monte, 28660 Madrid, Spain}

\begin{document}

\date{}

\maketitle

\begin{abstract}
    Computational fluid dynamics (CFD) provides high-fidelity simulations of fluid flows but remains computationally expensive for many-query applications. In recent years deep supervised learning (DL) has been used to construct data-driven fluid-dynamic surrogate models. In this work we consider a different learning paradigm and embrace generative modelling as a framework for constructing scalable fluid-dynamics surrogate models. We introduce {\it FluidFlow}, a generative model based on conditional flow-matching --a recent alternative to diffusion models that learns deterministic transport maps between noise and data distributions--. {\it FluidFlow} is specifically designed to operate directly on CFD data defined on both structured and unstructured meshes alike, without the needs to perform any mesh interpolation pre-processing and preserving geometric fidelity. We assess the capabilities of {\it FluidFlow}  using two different core neural network architectures --a U-Net and diffusion transformer (DiT)--, and condition their learning on physically meaningful parameters such as Mach number, angle of attack, or stagnation pressure --a proxy for Reynolds number--. The methodology is validated on two benchmark problems of increasing complexity: prediction of pressure coefficients along an airfoil boundary across different operating conditions, and prediction of pressure and friction coefficients over a full three-dimensional aircraft geometry discretized on a large unstructured mesh. In both cases, {\it FluidFlow} outperform strong multilayer perceptron baselines, achieving significantly lower error metrics and improved generalisation across operating conditions. Notably, the transformer-based architecture enables scalable learning on large unstructured datasets while maintaining high predictive accuracy. These results demonstrate that flow-matching generative models provide an effective and flexible framework for surrogate modelling in fluid dynamics, with potential for realistic engineering and scientific applications.
\end{abstract}

\section{Introduction}

Computational fluid dynamics (CFD) \cite{blazek2015computational} 
is the gold standard for the simulation of complex flow phenomena as well as the study of aerodynamic analysis and design, providing high-fidelity solutions 
of a range of partial differential equations that model the evolution of fluid flows, from Navier-Stokes equations to adequate approximations such as Reynolds-Averaged Navier-Stokes (RANS) or Large Eddy Simulations (LES).
However, its computational cost can become prohibitive when a large number of flow solutions is required, as in design-space exploration, uncertainty quantification, optimization, or real-time applications. For this reason, data-driven surrogate models have become increasingly attractive. Once trained, such models can provide approximate flow solutions at a cost that is orders of magnitude smaller than that of a full CFD simulation. Traditional surrogate modelling (e.g. proper orthogonal decomposition, dynamic mode decomposition, and others \cite{brunton2022data}) are increasingly superseded nowadays with the advent of modern machine learning methods \cite{Goodfellow-et-al-2016, brunton2020machine, le2023improving}. These include deep supervised learning approaches for the prediction of fluid-flow solutions and related properties via different deep learning arquitectures, such as deep multilayer perceptrons (MLPs), convolutional neural networks (CNNs), graph neural networks (GNNs) \cite{karniadakis2021physics,  pfaff2020learning, hines2023graph, elrefaie2024surrogate, ladron2025certifiable} or neural operators \cite{li2020fourier}; or via deep reinforcement learning paradigms \cite{ramos2025transfer}, among others.

\medskip In parallel, generative modelling (aka generative AI or genAI) \cite{banh2023generative} has had a spectacular growth and success in the last few years. Somewhat conceptually different from deep supervised learning, generative modelling aims to generate new, novel data instances that resemble the original training data. Accordingly, rather than learning a deterministic nonlinear input-output function as in supervised learning, generative models are trained to learn how the data is actually generated --i.e. learning a probability distribution over all plausible solutions--, thus allowing them to create entirely new, realistic samples at inference.\\
`Classical' generative modelling includes generative adversarial networks (GANs) \cite{goodfellow2014generative} or variational autoencoders (VAEs) \cite{kingma2013auto}, whereas more recent paradigms include autorregressive models \cite{vaswani2017attention}, and more lately diffusion models \cite{ho2020denoising, song2020score, rombach2022high} and flow-matching \cite{lipman2022flow}, all the latter having prompt-like inputs. Depending on their modelling choices and inductive biases, they have been successfully applied for text, image, audio, or video generation within traditional computer-science fields like natural language processing or computer vision, and some of these are currengly state of the art (SOTA) in their respective fields. For instance, popular image-generation software like DALL-E or Stable Diffusion are built on diffusion models (actually Stable Diffusion also uses VAEs), whereas popular large language models (ChatGPT, Claude) are built on autorregressive (transformer) generative models.
Of particular interest here is the class of {\it diffusion models} \cite{ho2020denoising, song2020score, rombach2022high}. Diffusion models generate data by gradually denoising a sample, initially starting from pure noise and iteratively reversing a forward process that corrupts data with Gaussian noise. In Denoising Diffusion Probabilistic Models (DDPM) \cite{ho2020denoising}, this is framed as a discrete-time Markov chain trained to predict and remove noise step-by-step, optimizing a variational objective. The Score-Based Generative Modeling through SDEs formulation \cite{song2020score} generalizes this to continuous time, learning the score function (gradient of the log-density) and generating samples by solving a reverse-time stochastic differential equation.\\
Similar in spirit to diffusion models is the concept of \emph{flow matching} \cite{lipman2022flow}. Like \cite{song2020score}, flow matching can also be viewed as a continuous-time alternative to classical denoising diffusion probabilistic models (DDPMs) \cite{ho2020denoising}, but instead of simulating a stochastic reverse-diffusion process, flow matching learns a {\it deterministic} vector field that transports samples from a simple reference distribution, typically Gaussian noise, to the target data distribution. In practice, this latter formulation is attractive because it simplifies both training and sampling, and often leads to lower computational cost than standard DDPM formulations while retaining the ability to model complex distributions \cite{wang2025fourierflow}.

\medskip
An intuitive application of generative modelling for scientific and engineering problems --including fluid dynamics-- is to serve as data augmentation techniques, e.g. for the construction of synthetic CFD datasets or for increasing resolution \cite{xie2018tempogan, molinaro2024generative}. Another promising application is the generation of novel designs conditioned on realistic aerodynamic constraints \cite{martin2025ai, graves2024airfoil}.
Interestingly, it is also possible to re-purpose generative modelling paradigms for {\it prediction}, i.e. for ML-based surrogate modelling traditionally approached through supervised learning.
For instance, denoising diffusion models have been explored for compressible fluid flow prediction \cite{abaidi2025exploring}, whereas
score-based diffusion models have been studied in aerodynamics to reproduce pressure fields and associated flow quantities with good accuracy, while also providing a probabilistic description of the solution \cite{liu2024uncertainty, wang2025aerodit, ogbuagu2026foildiff}. To the best of our knowledge, to date flow-matching models have been seldom applied in this context.\\
While re-purposing diffusion models for fluid-dynamics surrogate modelling is appealing, a crucial difficulty remains for many realistic applications: CFD data --needed to train the generative model-- are often defined on {\it unstructured} meshes. This is especially true for industrial geometries, where quantities such as pressure coefficient distributions are naturally available from irregular surface discretizations. At the same time, many of the diffusion model architectures proposed in the literature critically rely on convolutional neural networks (CNNs), which are best suited to data defined on structured, Cartesian (image-like) grids and are not designed to deal with unstructured ones. 
A possible workaround to this drawback is to preprocess the CFD data by interpolating the CFD-based flow fields onto a regular grid before training. Although practical, this step can discard part of the geometric information contained in the original mesh and introduce interpolation errors.

\medskip
Motivated by these facts, here we present {\it FluidFlow}: a family of flow-matching-based surrogate models for the {\it prediction} of fluid dynamics properties. {\it FluidFlow} can operate directly on CFD data in their native discretization, i.e. without the needs to pre-process the CFD data in any way, hence making it trainable on CFD data with both structured and unstructured meshes alike. We validate {\it FluidFlow} and assess its scalability to realistic problems by addressing a benchmark problem in fluid-dynamics surrogate modelling: (i) predicting the pressure coefficient ($C_p$) distribution of a 1D aerodynamic perimeter (airfoil boundary) and (ii) predicting aerodynamic (pressure and friction) coefficients on the surface of a three-dimensional full aircraft geometry. The ability of {\it FluidFlow} to accurately predict the whole distribution  of aerodynamic coefficients and to interpolate it across different physical parameters (Mach number ${M}_{\infty}$, far-field velocity $V_{\infty}$, angle of attack AoA, or stagnation pressure $p_i$ as a proxy for Reynolds number Re) is also assessed.
More concretely, for structured one-dimensional data --such as the pressure coefficient distribution along an airfoil surface-- we equip {\it FluidFlow} with both a  {\it U-Net} \cite{ronneberger2015u} or a {\it diffusion transformer} (DiT) \cite{peebles2023scalable} as two different core neural network architectures.
For the more challenging case of unstructured surface data defined on full (3D) aircraft configurations, {\it FluidFlow} mainly uses the diffusion transformer core, since its attention mechanism allows all points in the mesh to interact with each other without requiring a regular grid.
The prediction and interpolation capacity of our family of {\it FluidFlow} models across different operating conditions is made possible by directly conditioning these models on physically-meaningful parameters \cite{graves2024airfoil}, rather than on textual prompts or perceptual labels as generative modelling traditionally uses. 
This type of conditioning makes the framework particularly well-suited for predictive surrogate modelling, as the model learns not only to generate realistic aerodynamic flowfields, but those associated with the prescribed operating conditions.
Together, these two test cases allow us to assess both the accuracy of our approach and its ability to scale to realistic aerodynamic datasets.

\medskip \noindent 
The remainder of the paper is organized as follows. Section \ref{sec:methodology} presents in detail the methodology behind {\it FluidFlow}, including both datasets, the flow-matching formulation, classifier-free guidance \cite{ho2022classifier}, the two types of neural network architectures (U-Net and Diffusion transformer), and a concrete list of `tweaks' needed to put together {\it FluidFlow}, along some methods designed to scale-up models to large unstructured datasets.
Section \ref{sec:results} reports the results for both the airfoil (1D perimeter of an airfoil) and the aircraft (2D surface of a full 3D aircraft geometry) across a large range of operating conditions, concretely $(V_{\infty},\text{AoA})$ for the airfoil case and $(p_i,{M}_{\infty},\text{AoA})$ for the full 3D aircraft case. For the 1D (airfoil) case, we compare the generalisation performance of {\it FluidFlow} with the two neural network architectures (U-Net and DiT) against a vanilla MLP-type architecture that we train on the same task and dataset. We show that our flow-matching model --both using the U-Net and the DiT-- reach substantially smaller error metrics in the test set than the MLP, showing excellent results all over the airfoil, including regions with strong $C_p$ gradients, where classical ML-based modelling suffers. For the full (3D) aircraft geometry, {\it FluidFlow} results (with a DiT architecture) are still consistently superior to the SOTA MLP reported in the literature.
Finally, Section \ref{sec:conclusion} summarizes and discusses the main results and provides an outlook for possible future developments.

\section{Methodology}
\label{sec:methodology}
This section describes the datasets considered in this paper and the full generative modelling framework. For the sake of self-containedness, we provide a gently description of the flow-matching formulation before introducing the core neural network architectures.

\subsection{Datasets}

\subsubsection{Airfoil case: pressure coefficients over an airfoil's boundary}

The first test case concerns the generation of pressure coefficients $C_p$ at all spatial locations over an airfoil's boundary. To train our flow-matching model, we use the dataset reported in \cite{catalani2023comparative}, which contains the values of $C_p$ over an RAE2822 airfoil for 1000 
operating conditions, defined by combinations of 50 different Mach numbers (actually, the variable in the data is the far-field speed $V_{\infty}$, which itself is related to the Mach number) 
and 21 angles of attack $\text{AoA}$ and obtained by solving the RANS equations with the Spalart--Allmaras turbulence model at Reynolds number $\text{Re}= 6.5\times 10^6$. These two physical parameters (the variables upon which our flow-matching model will be conditioned) are grouped in the 2-dimensional vector 
$\mathbf{c} = (V_\infty,\text{AoA}) \in \mathbb{R}^{2}$.\\
The discretization of the airfoil's boundary is given in terms of a total of $N$ nodes, hence this boundary is fully parametrised by an $N$-dimensional vector $\mathbf{x}=(x_1,x_2,\dots,x_N)~\in~\mathbb{R}^{N}$, where $x_k$ denotes the pressure coefficient $C_p$ at the $k$-th spatial location of the airfoil. 
 In this work we may call each $\bf x$ associated to a given $\bf c$ a {\it sample}. 

\medskip \noindent 
The goal of our flow-matching model in this case is to learn from data the conditional distribution $p(\mathbf{x}\mid \mathbf{c})$, that is, the distribution of physically plausible $C_p$ `curves' associated with any given operating condition. The full dataset is split into a 70/15/15 partition for training, validation, and test. The split is only performed with respect to $\bf c$ (i.e., not in terms of $\bf x$), so as to assess the model's ability at inference to interpolate over operating conditions. The validation set is used mainly for selection of hyperparameters, see details in Appendix \ref{ap:hyper_airfoil}.

\subsubsection{Aircraft case: surface pressure coefficients over an unstructured mesh of the whole 3D geometry}

The second test case is substantially more challenging and involves the generation of the pressure coefficient $C_p$, along with the friction coefficient $C_f$, at all spatial locations on the surface of a full 3D aircraft geometry. Specifically, this corresponds to a four-dimensional vector associated with each spatial location, $(C_p, C_{f,x}, C_{f,y}, C_{f,z}) \in \mathbb{R}^4$.

\medskip \noindent 
We make use of the ONERA CRM WBPN dataset \cite{PETER2025106838}, which contains a total of 468 RANS simulations performed on the NASA/Boeing Common Research Model wing--body--pylon--nacelle configuration using the Spalart--Allmaras turbulence model. 
The simulations span different operating conditions, including 13 free-stream Mach numbers ($M_\infty$) ranging from 0.30 to 0.96. For each Mach number, 12 angles of attack (AoA) are considered, with the range narrowing as $M_\infty$ increases—specifically, from AoA $\in [-15^\circ, 15^\circ]$ at $M_\infty = 0.3$ and 0.5, down to AoA $\in [-8^\circ, 8^\circ]$ for $M_\infty$ between 0.88 and 0.96. To account for Reynolds number effects \cite{PETER2025106838} three stagnation conditions are included: a fixed stagnation temperature and three stagnation pressures $p_i = 10^5$, $2 \times 10^5$, and $4 \times 10^5$ Pa.
Accordingly, for this task and dataset, the flow-matching model will be conditioned on ${\bf c}=({M}_{\infty},\text{AoA},p_i)\in \mathbb{R}^3$.

\medskip \noindent 
Moreover, an unstructured mesh with a total of $N=260774$ spatial points covers the surface of the 3D geometry. Accordingly, the whole surface pressure/friction coefficients is summarised as a tensor (if spatial locations along the surface are parametrised in 2D), or as a matrix if we flatten the parametrisation of the $N$ spatial locations such that 
$
\mathbf{x} =(x_1,x_2,\dots,x_N)\in \mathbb{R}^{ 4 \times N}$, where $x_k=(C_p,\; C_{f,x},\; C_{f,y},\; C_{f,z})_k$ is the vector of pressure/friction coefficients at the $k$-th spatial location of the surface. 

\medskip \noindent As we shall see later, unlike the airfoil case, here there is no natural regular grid over which convolutions (natively defined within the U-Net architecture) could be applied directly. This will motivate the use of a transformer-based architecture rather than a U-Net, as the former can process the data as a sequence of points while still allowing to learn spatial correlations of pressure/friction coefficients through the attention mechanism. This issue is common in aerodynamic studies on unstructured domains, and its resolution is one of the central strengths of {\it FluidFlow}.\\
Here we use the train/test split provided with the database: 312 operating conditions for training and 156 for testing (i.e., like in the previous test case, we split the data with respect to the vector of operating conditions). To better compare our results against the benchmark, we keep the train/test split and thus don't perform hyperparameter optimisation in this case.


\subsection{Conditional flow matching}

We now describe the generative model used throughout this work. The central idea of flow matching is to learn how to transform samples from a simple distribution, usually Gaussian noise, into samples from the target CFD distribution.\\
Suppose that $\mathbf{x}$ is a CFD solution drawn from the data distribution and that $\boldsymbol{\varepsilon} \sim \mathcal{N}(\mathbf{0}, I)$ is a multivariate white Gaussian noise sample of the same dimension as $\mathbf{x}$. The flow-matching model employs a virtual time variable $t \in [0,1]$ which linearly interpolates the initial condition (Gaussian noise) at $t=0$ and the CFD solution $\bf x$ at time $t=1$. Accordingly, at a generic time $0<t<1$, the intermediate state ${\bf z}_t$ linearly interpolates the initial and final states
\begin{equation}
\mathbf{z}_t = (1-t)\boldsymbol{\varepsilon} + t\mathbf{x}.
\label{eq:linear_interpolation}    
\end{equation}
The time derivative of the interpolated state is a {\it velocity field}
\begin{equation}
\frac{d\mathbf{z}_t}{dt} = \mathbf{x} - \boldsymbol{\varepsilon}.\label{eq:velocity}    
\end{equation}
The role of the neural network inside the flow-matching model is to learn this velocity field. More precisely, the model learns the field
\begin{equation}
    \mathbf{v}_{\theta}(\mathbf{z}_t,t,\mathbf{c}),
    \label{eq:velocity_field}
\end{equation}
which takes as input the intermediate state $\mathbf{z}_t$, the time variable $t$, and the conditioning vector $\mathbf{c}$ \cite{graves2024airfoil}, and predicts the instantaneous velocity that moves the sample towards the target distribution.

\medskip \noindent 
We train the model using the conditional rectified flow-matching loss \cite{lipman2022flow}:
\begin{equation}
   \mathcal{L}_{\theta}
=
\mathbb{E}_{t,\mathbf{x},\mathbf{c},\boldsymbol{\varepsilon}}
\left[
\left\|
\mathbf{v}_{\theta}(\mathbf{z}_t,t,\mathbf{c}) - (\mathbf{x}-\boldsymbol{\varepsilon})
\right\|_2^2
\right],\label{eq:flow_matching_loss} 
\end{equation}
where $t \sim \mathcal{U}[0,1]$.
This loss has a simple interpretation: at randomly selected points along the path from noise to data, the network is asked to predict the correct direction of motion. Once the vector field has been learned, new samples can be generated by solving the ordinary differential equation
\begin{equation}
\label{eq:ODE_z}
\frac{d\mathbf{z}_t}{dt} = \mathbf{v}_{\theta}(\mathbf{z}_t,t,\mathbf{c}),
\end{equation}
starting from a random initial condition $\mathbf{z}_0 \sim \mathcal{N}(\mathbf{0},I)$ and integrating up to $t=1$. The final state $\mathbf{z}_1$ is then interpreted as the generated solution under the prescribed condition $\mathbf{c}$.

\medskip \noindent 
In the context of this paper, we refer to this generation procedure as \emph{sampling}. In practice, the ODE is solved numerically; in the present work, Euler integration is used for sampling.

\subsection{Classifier-free guidance}
During sampling, the quality of conditional generation can often be improved using a technique called classifier-free guidance (CFG) \cite{ho2022classifier}. Although this technique originated in the diffusion-model literature, its practical role can be adapted to the flow-matching formulation as follows.\\
The objective of conditional generation is to produce samples that are both (i) realistic and (ii) strongly consistent with the prescribed operating conditions. Now, these two objectives are not always perfectly aligned. A model may generate realistic fields that do not follow the condition closely enough, or it may overfit the condition at the expense of overall sample quality. CFG provides a simple way to tune this balance at inference time.
During training, the conditioning information of each sample is randomly removed with a probability of $p_{\text{drop}}=0.2$. In those cases, the model receives a learned `null condition', denoted by $\emptyset$, instead of the true condition vector $\bf c$. This enables the model to learn both a conditional prediction $\mathbf{v}_{\theta}(\mathbf{z}_t,t,\mathbf{c})$, and an unconditional one $\mathbf{v}_{\theta}(\mathbf{z}_t,t,\emptyset)$. Once trained, 
at sampling (inference) both predictions are combined as
\begin{equation}
\mathbf{v}_{\theta}(\mathbf{z}_t,t,\mathbf{c})_{\mathrm{CFG}}
=
\mathbf{v}_{\theta}(\mathbf{z}_t,t,\emptyset)
+
s
\left[
\mathbf{v}_{\theta}(\mathbf{z}_t,t,\mathbf{c})
-
\mathbf{v}_{\theta}(\mathbf{z}_t,t,\emptyset)
\right],
\label{eq:CFG}    
\end{equation}
where $s$ is the so-called guidance scale. When $s=1$, the expression reduces to standard conditional sampling. Larger values of $s$ increase the influence of the prescribed condition, although excessively large values may degrade sample quality.

\subsection{Neural network architectures}
As explained before, flow-matching formulation requires the need to define a neural network architecture to learn the velocity field (Eq.~\ref{eq:velocity_field}).
Two neural-network architectures are considered in this work: a U-Net \cite{ronneberger2015u} and a diffusion transformer (DiT) \cite{peebles2023scalableDiT}. The choice between them is dictated mainly by the structure of the data.
Concretely, the U-Net is solely used for the airfoil dataset, where the target variable is a one-dimensional signal defined on an ordered set of points. In this setting, local correlations along the surface can be exploited through one-dimensional convolutions. On the other hand,
the DiT is also used both for the airfoil case and, more importantly, for the aircraft dataset. Its key advantage is that it does not require the elements of $\bf x$ to lie on a regular grid. Instead, {\bf x} is treated by DiT as a sequence of points or patches, and the attention mechanism allows each element $x_k$ in $\bf x$ to `see' and `exchange' information with all others, thereby enabling learning spatial correlations which typically emerge in aerodynamic data {without the need to interpolate each $x_k$ into a structured mesh to use convolutions}. 

\subsubsection{Conditioning and time embeddings}
Both neural architectures --described in detail in the next sections-- are conditioned in the same general way. The flow parameters, such as Mach number ${M}_{\infty}$ and the angle of attack $\text{AoA}$ are scalar inputs. Rather than injecting these scalars directly into the network, we first map them into a higher-dimensional embedding space using a two-layer multilayer perceptron (MLP). This is standard practice in conditional generative modeling, because it gives the network more expressive features with which to represent the condition.

\medskip \noindent 
The time variable $t$ in the flow-matching ODE (Eq.~\ref{eq:ODE_z}) is embedded in a similar fashion. First, a frequency-based positional encoding is applied to $t$, and the resulting vector is passed through a two-layer MLP, following the strategy introduced in diffusion models \cite{ho2020denoising}. Intuitively, this allows the network to distinguish early, intermediate, and late stages of the transport process more effectively than if the scalar value $t$ was used directly.

\medskip \noindent 
Both the condition and the time embeddings are then combined and used to modulate intermediate activations via adaptive Layer Normalization Zero (adaLN-Zero), following \cite{peebles2023scalableDiT}. In this formulation, an auxiliary MLP predicts three sets of coefficients: a shift $\beta$, a scale $\gamma$, and a gate $\alpha$. If $l(\cdot)$ denotes a neural-network layer, the conditioned output can be written schematically as
\begin{equation}
   \alpha \, l\!\left(x(1+\gamma)+\beta\right). 
   \label{eq:embeddings}
\end{equation}
This mechanism provides a convenient and effective way to inject physical operating conditions into deep architectures.

\begin{figure}[htbp]
 \centering

 \begin{subfigure}[b]{0.35\textwidth}
 \centering
 \includegraphics[width=\textwidth]{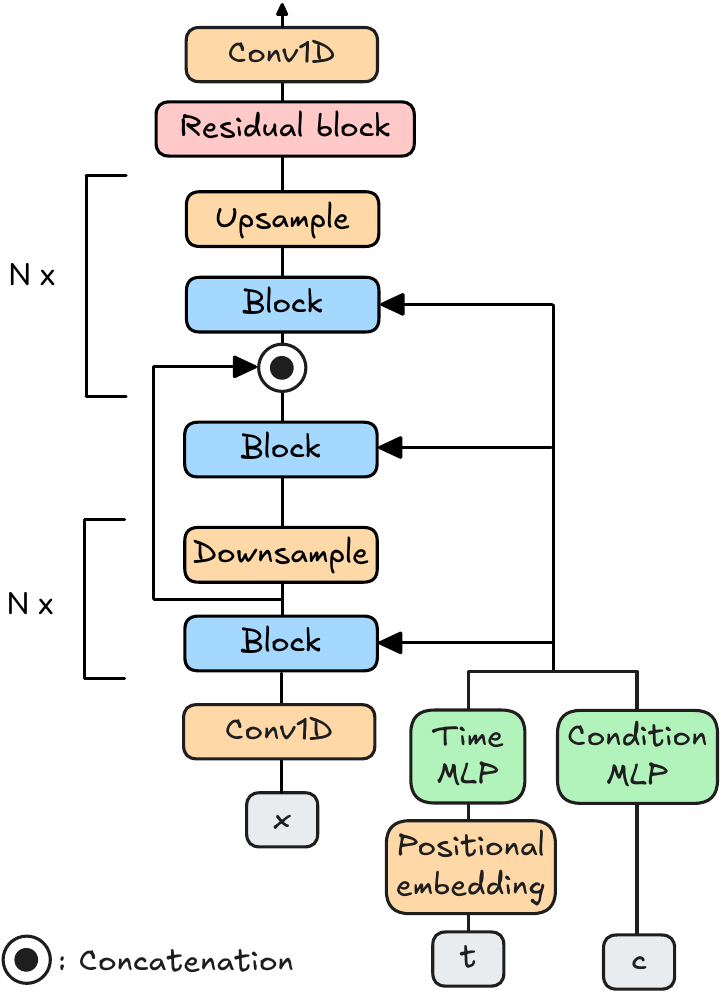}
 \caption{U-Net architecture.}
 \label{fig:unet}
 \end{subfigure}
 \begin{subfigure}[b]{0.25\textwidth}
 \centering
 \includegraphics[width=\textwidth]{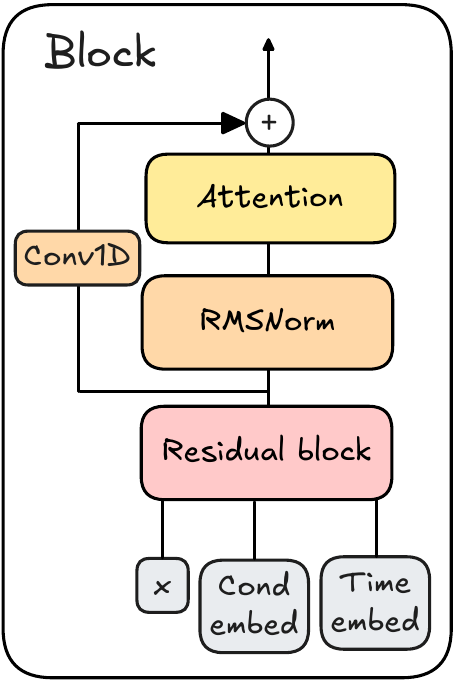}
 \caption{Basic block.}
 \label{fig:basic_block}
 \end{subfigure}
 \begin{subfigure}[b]{0.35\textwidth}
 \centering
 \includegraphics[width=\textwidth]{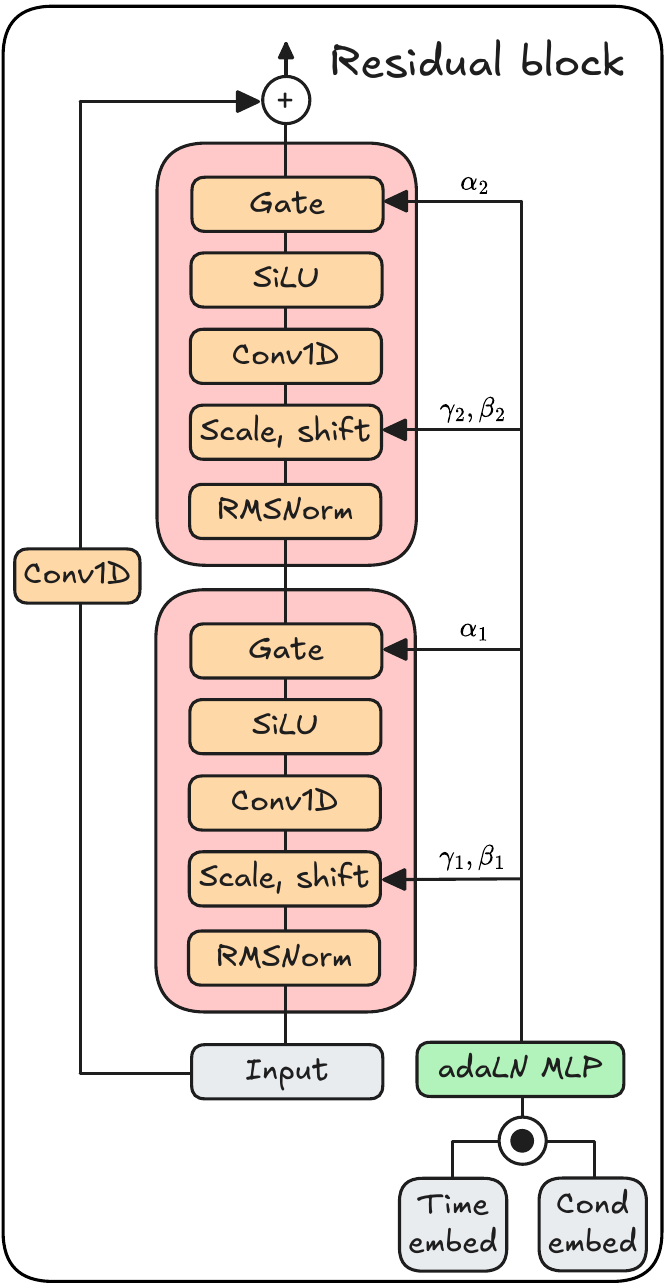}
 \caption{Residual block.}
 \label{fig:residual_block}
 \end{subfigure}

 \caption{U-Net architecture and its main components.}
 \label{fig:unet_architectures}
\end{figure}

\subsubsection{U-Net}

The U-Net used in this work is suitably adapted from the architecture commonly employed in diffusion models \cite{ho2020denoising}. Since the airfoil data are one-dimensional, all convolutions are one-dimensional as well.
The overall structure of the U-Net is shown in Fig.~\ref{fig:unet_architectures}. As in the classical U-Net design, the input passes through an encoder--decoder structure with skip connections between corresponding resolution levels. This architecture combines two useful properties. First, convolutions capture local patterns, which in this case correspond to variations of the pressure coefficient along neighboring points of the airfoil boundary. Second, attention blocks are included to enrich the representation with more global information, allowing the model to connect distant parts of the profile when necessary.

\medskip \noindent 
The detailed building blocks are also shown in Fig.~\ref{fig:unet_architectures}. Each block contains a conditioned residual unit and an attention mechanism. Root mean square normalization (RMSNorm) is used in place of standard normalization layers, and conditioning is injected through the adaLN-Zero mechanism described in the previous section.

\subsubsection{Diffusion transformer (DiT)}
The second neural network architecture we consider here is a diffusion transformer, based on \cite{peebles2023scalableDiT}, albeit with some important modifications relative to the original DiT formulation, that we outline below.

\medskip \noindent 
First, standard normalization layers are replaced by RMSNorm \cite{zhang2019root}. Second, the pointwise feedforward network is replaced by a SwiGLU block \cite{shazeer2020glu}. These changes are motivated by recent empirical evidence showing improved performance in related generative architectures \cite{yao2025reconstruction}. Third, because our data are not images, the original two-dimensional patchification stage is replaced by a one-dimensional version {(see Fig.~\ref{fig:patching} for an illustration).\\
This last point is particularly important. Observe that in a traditional image-generation setting, the transformer receives a sequence of image patches extracted from a two-dimensional array. In our case, the input $\bf x$ is instead either a one-dimensional sequence of points (the pressure coefficients over the airfoil's boundary) or a sequence of points on an unstructured mesh. The {\it patchify} layer therefore groups neighboring entries in the sequence into tokens of fixed size. If the input contains $N$ points and the patch size is $p$, the patchify operation produces $N/p$ tokens of dimension $d$, where $d$ is the transformer embedding dimension.

\begin{figure}[htbp]
 \centering
 \includegraphics[width=\textwidth]{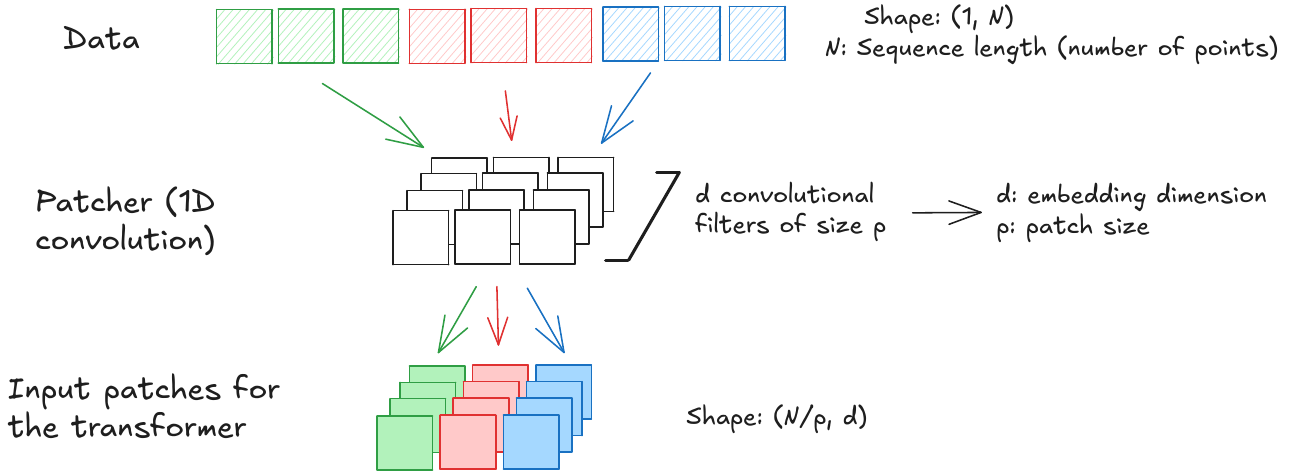}
 \caption{One-dimensional patchification used in the diffusion transformer.}
 \label{fig:patching}
\end{figure}

\medskip \noindent
An important practical consequence is that the patch size controls the sequence length seen by the transformer. Larger patch sizes reduce computational cost because fewer tokens are processed, but they also coarsen the representation of the flow field. As shown later in the results, this trade-off is particularly relevant for aerodynamic data.

\subsection{Scaling to large unstructured datasets: linear attention}
As mentioned previously, the aircraft dataset, coming from unstructured mesh data, cannot be handled by a U-Net architecture, as its internal convolutions assume an underlying local ordering that is only meaningful in a structured sense. For this reason, the aircraft case is handled exclusively through a flow-matching model with a diffusion transformer architecture.\\
Now, the aircraft dataset poses a much more demanding problem than the airfoil case. Not only are there more output channels (i.e. at each spatial location the output is a four-dimensional vector $(C_p,\; C_{f,x},\; C_{f,y},\; C_{f,z}) \in \mathbb{R}^4$, instead of a scalar $C_p \in \mathbb{R}$), but each sample in the aircraft dataset contains more than $2.6\times 10^5$ mesh points. 
Even for transformers, the computational burden is severe. For instance, standard multi-head self-attention has quadratic complexity $O(N^2)$ with respect to the sequence length $N$. For meshes of this size, the memory and time requirements become very large. To make training feasible, we have performed the following modifications: (i) first, we compile the model; (ii) second, we adopt mixed-precision training with \texttt{bfloat16}, where queries and keys are normalized to ensure training stability following \cite{dehghani2023scaling}. Finally, (iii) we investigate a lower-cost attention approximation based on linear attention \cite{katharopoulos2020transformers}. Linear attention reduces the $O(N^2)$ complexity by replacing the softmax kernel with a feature map that allows the matrix multiplications to be reordered.
Following the strategy used in Sana \cite{xie2024sana}, we use ReLU as the feature map. The attention operator is formally written as
\begin{equation}
\label{eq:linear_attn}
\text{LinearAttention}(Q,K,V)
=
\frac{
\text{ReLU}(Q)\left(\text{ReLU}(K)^T V\right)
}{
\text{ReLU}(Q)\left(\sum_j \text{ReLU}(k_j)^T\right)
}.
\end{equation}
where $Q$, $K$ and $V$ are the so-called Query, Key, and Value matrices in `attention jargon': learned projections of the input that let the model decide what to look at, where to look, and what information to extract.
The essential idea is that the product between keys and values is computed first, yielding a global context matrix. This allows the total cost to scale linearly with the sequence length rather than quadratically. The approximation is therefore attractive for very long sequences such as those encountered in unstructured CFD datasets, although some loss in predictive accuracy may occur.

\section{Results}
\label{sec:results}
To train all flow-matching models in {\it FluidFlow} the entries of $\bf c$ and the target variables $C_p$ and $C_f$ are standardized, although all the metrics have been computed on their original scale. This section reports the predictive performance on the test sets for the proposed models on the two datasets considered above. Unless otherwise stated, all samples are generated using Euler integration for Eq.~\ref{eq:ODE_z} with 500 steps, and classifier-free guidance. For additional details and video visualisations of the flow-matching trajectories for both cases under study, see \cite{miniweb}.


\subsection{Airfoil case}
On top of our two flow-matching models, we also trained for comparison a multilayer perception (MLP) on the same dataset. Details on the MLP, its hyperparameter optimisation, and the hyperparameter optimisation of the two flow-matching models are available in Appendix \ref{ap:hyper_airfoil}.
To assess the performance of the two {\it FluidFlow} models and the MLP in the test set, we consider a set of global error metrics that includes (i) mean square error (MSE), root-mean square error (RMSE), mean abolute error (MAE) and mean relative error (MRE) (all of them averaged over all spatial locations and for all operating conditions in the test set), (ii) 
the 95 and 99 percentiles of the absolute error distribution of $|C_p - \hat{C}_p|$ over spatial locations and operating conditions, (iii) the coefficient of determination $R^2$ from a scatter plot between the ground true $C_p$ and the prediction $\hat{C}_p$ (for all spatial locations and operating conditions in the test set) and (iv) a relative $L^2$ error of the output vector resulting of stacking the prediction for all points in the test set $\lVert \mathbf{y}_{\text{true}} - \mathbf{\hat{y}} \rVert_2/\lVert \mathbf{y}_{\text{true}} \rVert_2$. 
Table \ref{tab:model_comparison_airfoil} summarises these metrics and compares these for our two flow-matching models (the one with a U-Net and the one with a DiT) and for the MLP we independently trained on the same dataset.
The results show a clear improvement of both {\it FluidFlow} models over the MLP baseline. Both the U-Net and the DiT versions of {\it FluidFlow} reduce substantially all error metrics of the MLP baseline and achieve excellent $R^2$ values. For this structured one-dimensional problem, the U-Net and the DiT perform similarly, which suggests that both local convolutional processing and global attention-based processing are effective when the input geometry is simple. At this point we should mention that authors in \cite{catalani2023comparative} also build a prediction pipeline based on convolutional neural networks (CNN) and geodesic convolutional neural networks (GCNN). We couldn't really compare our results against that benchmark because their performance results --in terms of absolute errors $|C_p - \hat{C}_p|$-- were disaggregated and only heatmaps for different Mach conditions and a few angles of attack were presented, i.e. no global MAE was constructed. From the range of those heatmap legends it seems that our results remain quite competitive as compared to CNN and GCNN, but a more careful comparison is needed.

\begin{table}[htbp]
\centering
\begin{tabular}{lcccccccc}
\hline
Model & mse & rmse & mae & mre (\%) & ae\_95 & ae\_99 & $R^2$ & Relative $L^2$ \\
\hline
Vanilla MLP & 0.00129 & 0.03598 & 0.01763 & 16.85219 & 0.05716 & 0.14176 & 0.99730 & 0.04911 \\
{\it FluidFlow} (U-Net) & 0.00009 & 0.00961 & 0.00240 & 4.48810 & 0.00761 & 0.03175 & 0.99981 & 0.01325 \\
{\it FluidFlow} (DiT) & 0.00009 & 0.00953 & 0.00249 & 3.43723 & 0.00764 & 0.03246 & 0.99981 & 0.01314 \\
\hline
\end{tabular}
\caption{Generalisation performance comparison for the airfoil task.}
\label{tab:model_comparison_airfoil}
\end{table}

\medskip \noindent 
For a better characterisation of where errors lie \cite{ESWA}, in Figure \ref{fig:scatters_airfoil} we report scatter plots of the true vs predicted pressure coefficient, for all locations and operating conditions in the test set (we have a total of 152 different operating conditions in the test set, and each case has 691 spatial locations, hence scatter plots have 105032 points). We clearly see that matching between prediction and ground truth is excellent and consistently better for the flow-matching models, where only small errors emerge close to $C_p\approx -1$. 

\begin{figure}[htbp]
 \centering
 \includegraphics[width=1.\textwidth]{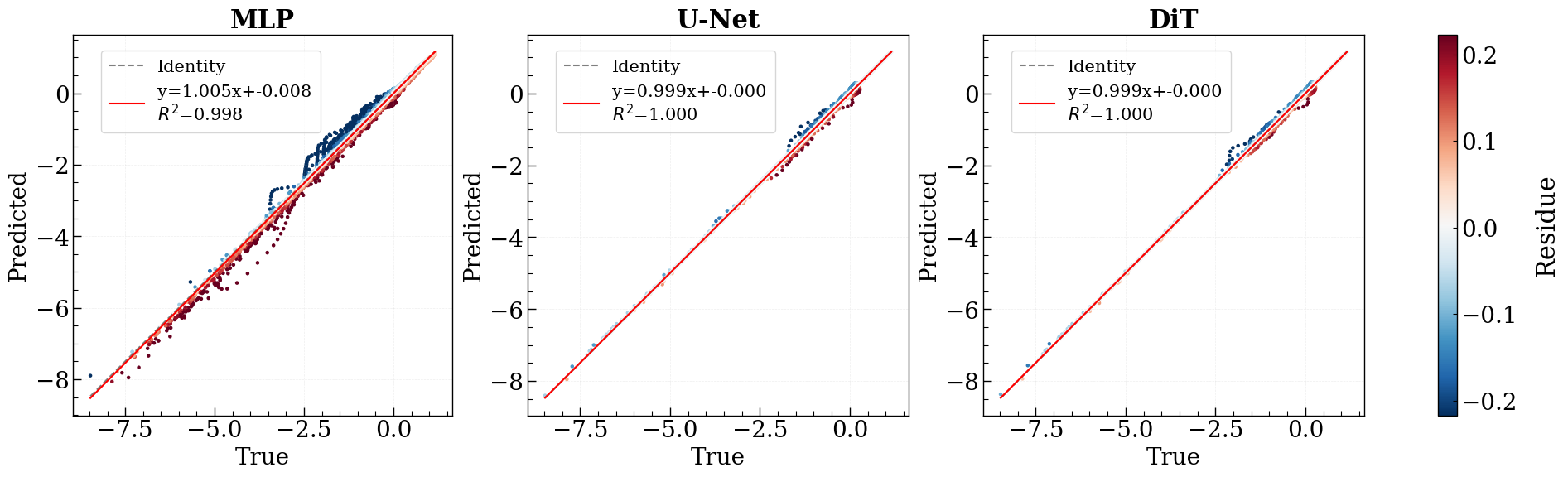}
 \caption{Scatter plots of the true vs predicted pressure coefficient, for all locations and operating conditions in the airfoil's test set. Color codes the magnitude of the residuals.}
 \label{fig:scatters_airfoil}
\end{figure}

Figure \ref{fig:airfoils_unet} provides representative examples of predicted and reference pressure distributions for four different operating conditions. The agreement is generally very good over the full airfoil surface, including the regions of strong pressure variation, where MLP struggles to produce accurate results. This is encouraging because it indicates that the model is not merely fitting smooth trends, but is also reproducing the more sensitive features of the aerodynamic loading.

\begin{figure}[htb!]
 \centering
 \includegraphics[width=0.8\textwidth]{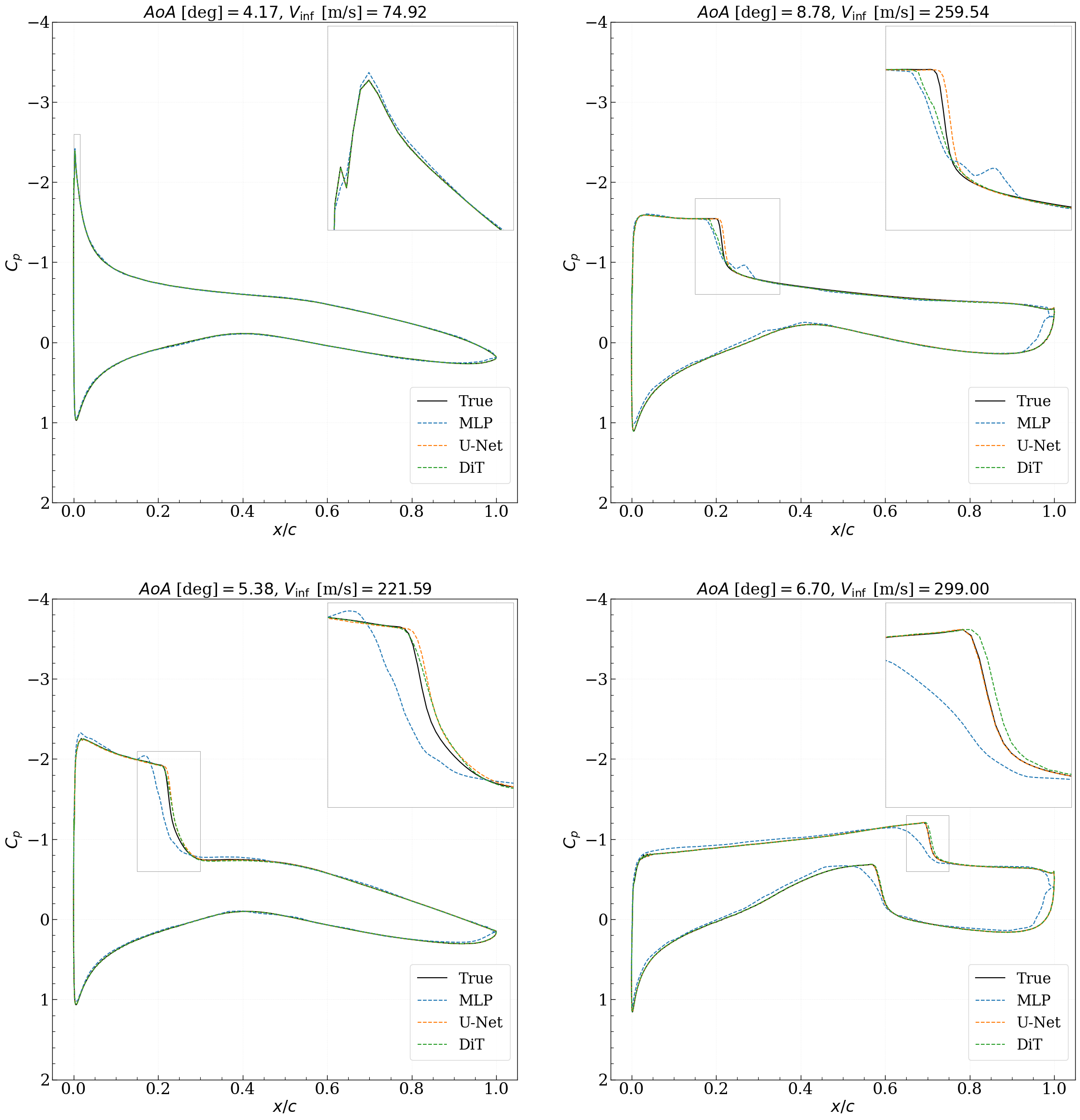}
 \caption{Examples of predicted and reference airfoil pressure distributions for different operating conditions.}
 \label{fig:airfoils_unet}
\end{figure}

\medskip
We finally checked that the value of the guidance scale $s$ of the classifier-free guidance protocol (Eq.~\ref{eq:CFG}) did not have a major impact on test metrics, see Table \ref{tab:cfg_comparison} for a comparison of error metrics for the DiT-based flow matching model with different values of $s$. 

\begin{table}[ht]
\centering
\begin{tabular}{lcccccccc}
\toprule
$s$ & mse & rmse & mae & mre (\%) & ae\_95 & ae\_99 & $R^2$ & Relative $L^2$ \\
\midrule
1.0 & 0.000110 & 0.0105 & 0.00289 & 3.520 & 0.00920 & 0.0360 & 0.99977 & 0.0145 \\
1.5 & 0.000092 & 0.0096 & 0.00252 & 3.089 & 0.00766 & 0.0339 & 0.99980 & 0.0132 \\
2.0 & 0.000091 & 0.0095 & 0.00249 & 3.043 & 0.00764 & 0.0325 & 0.99981 & 0.0131 \\
2.5 & 0.000094 & 0.0097 & 0.00254 & 3.118 & 0.00793 & 0.0321 & 0.99980 & 0.0134 \\
3.0 & 0.000100 & 0.0100 & 0.00261 & 3.205 & 0.00824 & 0.0319 & 0.99979 & 0.0138 \\
4.0 & 0.000113 & 0.0106 & 0.00278 & 3.393 & 0.00900 & 0.0306 & 0.99976 & 0.0146 \\
6.0 & 0.000162 & 0.0127 & 0.00396 & 4.656 & 0.01325 & 0.0365 & 0.99966 & 0.0176 \\
\bottomrule
\end{tabular}
\caption{DiT-based flow-matching model performance metrics across different CFG guidance scales $s$, for the airfoil case, showing minor dependence on $s$.}
\label{tab:cfg_comparison}
\end{table}

\subsection{Aircraft case}

Since the pressure coefficients in the ONERA CRM WBPN dataset are defined on the nodes of a large unstructured mesh, {\it FluidFlow} uses here only the diffusion transformer, with patch size equal to 1 and linear attention (the whole model configuration is defined in Appendix \ref{ap:hyper_aircraft}). 
Table \ref{tab:r2_comparison_onera} compares the performance of {\it FluidFlow} against the best baseline reported in the reference study \cite{PETER2025106838}. For consistency with that work, the reported $R^2$ values are computed using their weighted definition:
\begin{equation}
 R_y^2 = 1 - \frac{\sum_{i=0}^{n_p \times n_{te}} w_f (y_i - \hat{y}_i)^2}
 {\sum_{i=0}^{n_p \times n_{te}} w_f (y_i - \bar{y})^2},
 \qquad
 y \in \{C_p, C_{f,x}, C_{f,y}, C_{f,z}\},
 \label{eq:weighted_r2}
\end{equation}
where $w_f$ is a flow-dependent weighting factor and $\bar{y}$ is the mean value of the variable over all spatial locations and test conditions. In the reference protocol, $w_f=1$ for cases with angle of attack in the interval $[-10^\circ,10^\circ]$, while cases outside that range receive a weight of $0.5$ to reduce the influence of simulations expected to be less reliable.

\begin{table}[htbp]
\centering
\begin{tabular}{lccccc}
\hline
Model & $R^2$ & $R^2_{C_p}$ & $R^2_{C_{f,x}}$ & $R^2_{C_{f,y}}$ & $R^2_{C_{f,z}}$ \\
\hline
MLP \cite{PETER2025106838} & 0.956 & 0.972 & 0.944 & 0.951 & 0.957 \\
{\it FluidFlow} (DiT) & 0.965 & 0.974 & 0.959 & 0.960 & 0.965 \\
\hline
\end{tabular}
\caption{Comparison of the performance of {\it FluidFlow} with the baseline MLP model \cite{PETER2025106838} for the 3D aircraft task.}
\label{tab:r2_comparison_onera}
\end{table}

\noindent {\it FluidFlow} improves the global weighted SOTA $R^2$ from 0.956 to 0.965, with gains across all predicted variables. Although the improvement may appear moderate in absolute terms, it is obtained on a highly demanding high-dimensional regression task involving a very large unstructured mesh.

\medskip \noindent 
For completeness, Table \ref{tab:performance_metrics_dit} reports additional error metrics computed with the standard, unweighted formulation alongside other quantifiers defined in the previous section. These values provide a more conventional view of the predictive error of the model.

\begin{table}[htbp]
\centering
\begin{tabular}{ccccccccc}
\hline
MSE & RMSE & MAE & MRE & ae\_95 & ae\_99 & $R^2$ & Relative $L^2$ \\
\hline
0.0013 & 0.0358 & 0.0065 & 23.4211\% & 0.0271 & 0.1229 & 0.9659 & 0.1835 \\
\hline
\end{tabular}
\caption{Additional performance metrics for {\it FluidFlow} on the 3D aircraft case in the ONERA CRM WBPN dataset (full 3D geometry). 
}
\label{tab:performance_metrics_dit}
\end{table}

\noindent 
Finally, in Figure \ref{fig:ejemplo_onera} we plot heatmaps of the  CFD-generated pressure/friction fields (top panels) and the predictions of {\it FluidFlow} (bottom panels) over the whole surface, for the operating condition $
M_\infty = 0.3, \  \text{AoA} = -6^\circ, \  p_i = 10^5$.
The generated (predicted) fields reproduce with notable accuracy all the nuanced spatial patterns, indicating that the model captures the dominant aerodynamic structures despite the complexity of the geometry and the size of the mesh. For video visualisations of the generated fields, see \cite{miniweb}. 

\begin{figure}[htb!]
 \centering
 \includegraphics[width=\textwidth]{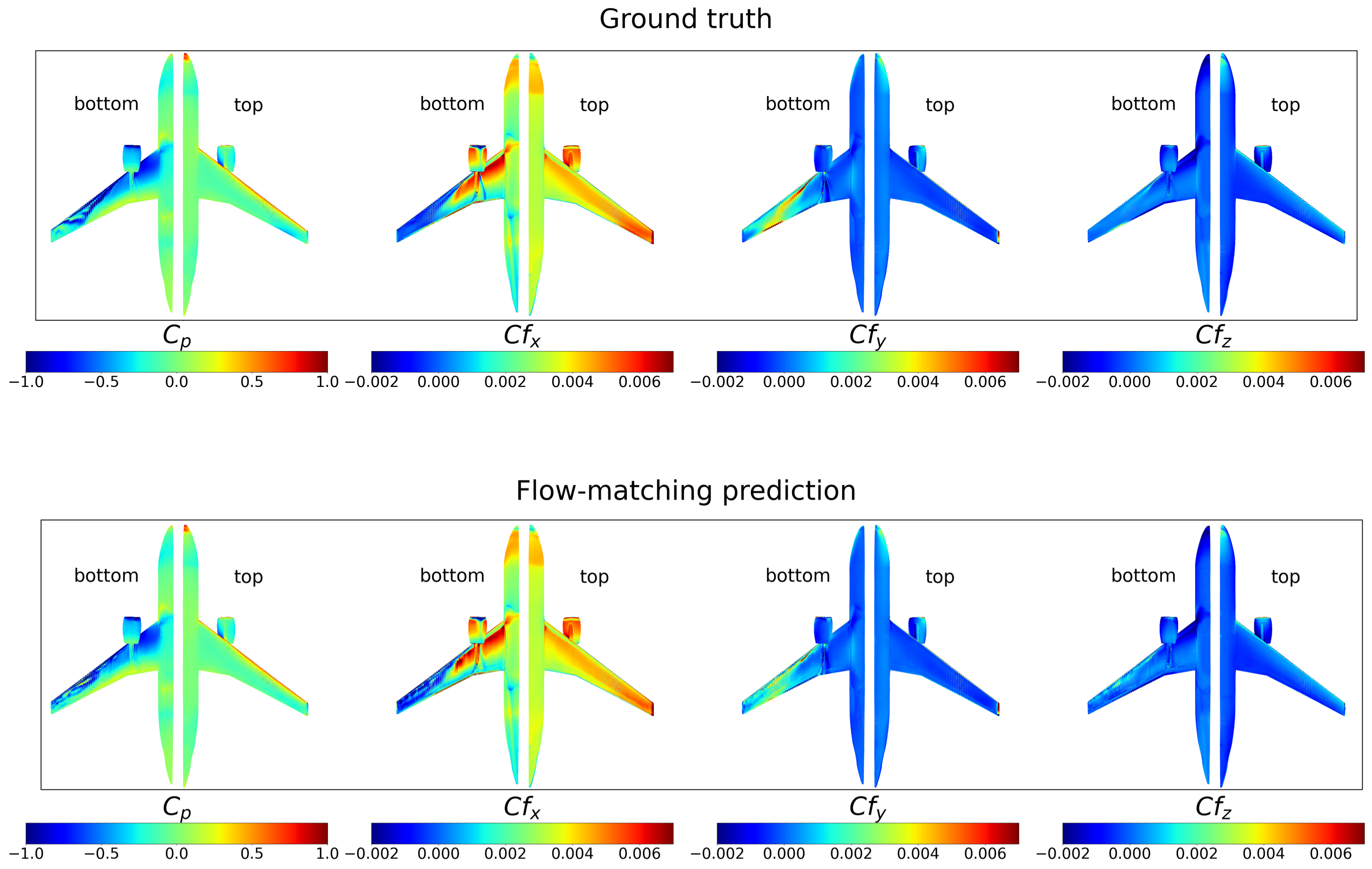}
 \caption{Comparison between (ground true) CFD pressure/friction coefficient fields (top panels) and the prediction generated by the DiT flow-matching model (bottom panels) for one particular operating condition with parameters $p_i=1\times10^5, M_{\infty}=0.3$ and $\text{AoA}=-6º$.} 
 \label{fig:ejemplo_onera}
\end{figure}

\section{Conclusions}
\label{sec:conclusion}
This work presents {\it FluidFlow}, a generative surrogate modelling framework for fluid-dynamic problems based on a suitable repurposing of conditional flow-matching. {Our pipeline is designed to equally work with CFD data on structured and unstructed meshes, hence enabling its application to realistic problems of scientific, engineering and industrial interest.}
The approach has been evaluated on two aerodynamic problems of increasing complexity: the prediction of pressure coefficients on an airfoil across different operating conditions, and the prediction of both pressure and friction coefficients over the complete surface of a 3D realistic aircraft geometry
(with spatial locations provided by a large unstructured aircraft mesh), over different operating conditions.\\
For the airfoil dataset, flow-matching models with both the U-Net and the diffusion transformer architectures perform similarly, and both clearly outperform a standard MLP baseline. This result shows that flow-matching-based generative models are competitive even in relatively simple structured settings, where deterministic regressors are already very strong.\\
A significant result is also found for the full (3D) aircraft dataset. In this case, the data are defined on a large unstructured mesh, which makes conventional convolution-based approaches difficult to apply directly. By using a flow-matching model with an adequately adapted transformer architecture, {\it FluidFlow} can direcly operate on the native discretization of the CFD data, avoiding interpolation pre-processing of the data onto a structured grid. The resulting predictions improve upon the baseline model reported for the same dataset, which suggests that generative formulations are a promising direction for aerodynamic surrogate modeling beyond simplified benchmark cases.

\medskip \noindent 
From the broader perspective of fluid-mechanical surrogate modelling, the main message from this work is that generative modelling --concretely, flow-matching-- should not be viewed only as machine-learning curiosities inherited from sophisticated image generation frameworks. On the contrary, they provide a practical framework for learning families of flow solutions conditioned on physically meaningful parameters, while remaining flexible enough to handle complex geometries and irregular discretizations.\\
Several directions remain open for future work. On a technical level, the trade-off between patch size, attention mechanism, and predictive fidelity should be studied more systematically, especially for very large meshes. Likewise, the effect of linear attention on accuracy and computational cost deserves a dedicated analysis. Finally, improved parallelization strategies for multi-GPU settings require further development, especially to effectively address larger-scale problems.
In more general terms, the present work has focused on steady aerodynamic quantities, so extending {\it FluidFlow} to non-stationary three dimensional flows would be a natural next step. Its potential to not only accurately interpolate over operating conditions but also over different 3D geometries --i.e. eventually reaching foundation-model status \cite{lu2021learning, li2023fourier, azizzadenesheli2024neural}-- is a challenging yet exciting open question. Finally, in this work we focus on fluid-dynamical surrogates, but the pipeline could be extended to many other problems 
in science and engineering where one needs to integrate some PDE over a complex geometry. Examples range from structural mechanics (solid mechanics, elasticity, fracture, crash simulation), electromagnetism (Maxwell’s equations are solved in irregular domains such as antennas or waveguides), seismic wave propagation in geophysics, edge plasma modeling in Tokamaks and many others.\\

\noindent {\bf Acknowledgments --} 
The authors acknowledge funding from project TIFON (PLEC2023-010251) funded by
MCIN/AEI/10.13039/501100011033, Spain. 
This research has received funding from the European Union (ROSAS, project number 101138319). 
This research has received funding from the European Research Council Executive Agency under TRANSDIFFUSE project, Grant Agreement No. 101167322.
Views and opinions expressed are, however, those of the authors only and do not necessarily reflect those of the European Union. Neither the European Union nor the granting authority can be held responsible for them.
LL acknowledges partial support from project CSxAI (PID2024-157526NB-I00) funded by MICIU/AEI/10.13039/501100011033/FEDER, UE, project Maria de Maeztu (CEX2021-001164-M) funded by the MICIU/AEI/10.13039/501100011033, and from the European Commission Chips Joint Undertaking project No. 101194363 (NEHIL).
GR acknowledges partial financial support received by the Grant DeepCFD (Project No. PID2022-137899OB-I00) funded by MICIU/AEI/10.13039/501100011033 and by ERDF, EU. 
This work was supported by the EuroHPC Joint Undertaking through access to the MareNostrum5 supercomputer at the Barcelona Supercomputing Center (BSC), Spain, under the EuroHPC Benchmark Access Call EHPC-BEN-2026B03-071. 
Finally, all authors gratefully acknowledge the Universidad Politécnica de Madrid for providing computing resources on Magerit Supercomputer.

\medskip \noindent {\bf Code and visualisations --} Codes are available at \cite{code}, and other details, videos and visualisations are available at \cite{miniweb}.


\appendix

\section{Specifications and hyperparamenter optimisation for the three models on the 1D airfoil dataset}
\label{ap:hyper_airfoil}

{{\bf Baseline MLP --}  The baseline model used for comparison is a standard fully-connected multilayer perceptron (MLP) with a uniform architecture, i.e., all hidden layers share the same number of neurons for simplicity. The network operates in a pointwise manner: for each query location, the input is defined by the spatial coordinates of the point at which the pressure coefficient is to be predicted, together with the corresponding operational flow parameters. The model is trained using the Adam optimizer, minimizing the mean squared error (MSE) loss function under a mini-batch training strategy. A step-based learning rate scheduler is employed, with a decay applied every epoch. The activation function used across all hidden layers is the Exponential Linear Unit (ELU), and the network weights are initialized using the Xavier uniform initialization scheme.\\
Hyperparameter tuning is conducted using the validation dataset and the Optuna framework~\cite{optuna_2019}. The procedure consists of defining search spaces for the relevant hyperparameters and exploring them through the Tree-structured Parzen Estimator (TPE) sampler~\cite{watanabe2023tree}. For each sampled configuration, the model is trained on the training set and evaluated on the validation set using the MSE metric. The optimal configuration is selected as the one that yields the lowest validation error. The set of tuned hyperparameters includes the number of hidden layers, the number of neurons per layer, the batch size, the learning rate, the probability of dropout, and the total number of training epochs (see \autoref{tab:mlp_airfoil_params} for a detailed summary). In total, $200$ hyperparameter configurations were evaluated during the optimization process.}

\begin{table}[h]
\centering
\begin{tabular}{ll}
\toprule
\textbf{Parameter} & \textbf{Value} \\ 
\midrule
Optimizer & Adam \\
Loss function & Mean Squared Error (MSE) \\
Activation function & ELU \\
Weight initialization & Xavier uniform \\
Initial learning rate & $1.47 \times 10^{-3}$ \\
Learning rate scheduler & Step scheduler (step size = 1 epoch) \\
Learning rate decay factor & 0.99 \\
Batch size & $177$ \\
Hidden layer dimension & $113$ \\
Number of hidden layers & $10$ \\
Dropout probability & $7.76 \times 10^{-4}$ \\
Number of training epochs & $58$ \\
Number of hyperparameter trials & $200$ \\
\bottomrule
\end{tabular}
\caption{Summary of the MLP architecture, training configuration, and optimized hyperparameters.}
\label{tab:mlp_airfoil_params}
\end{table}

\medskip \noindent {\bf Selection of hyperparameters for the flow-matching models --} The selection of hyperparameters we used for both the U-Net and DiT architectures of the flow-matching models on the airfoil dataset are reported in Tables  \ref{tab:unet_airfoil_params} and \ref{tab:dit_airfoil_params} respectively. During training, a cosine learning rate scheduler (pytorch's CosineAnnealingLR) with no warmup has been used, with a final learning rate of $10^{-6}$.
\begin{table}[h]
\centering
\begin{tabular}{ll}
\toprule
\textbf{Parameter} & \textbf{Value} \\ 
\midrule
Blocks dim & [128, 256, 512] \\
Num. attention heads & 8 \\
Attention hidden dim & 512 \\
$p_{drop}$ & 0.2 \\
Total parameters & 43.2M \\
Training steps & 200000 \\
Batch size & 64 \\
Optimizer & AdamW \\
Learning rate & $2 \times 10^{-4}$ \\
Weight decay & $10^{-2}$ \\
Max gradient norm & 1.0 \\
Floating point precision & float32 \\
\bottomrule
\end{tabular}
\caption{Neural network and training hyperparameters of the U-Net architecture on the airfoil case.}
\label{tab:unet_airfoil_params}
\end{table}

\begin{table}[h]
\centering
\begin{tabular}{ll}
\toprule
\textbf{Parameter} & \textbf{Value} \\ 
\midrule
Num. blocks & 6 \\
Num. heads & 4 \\
Hidden dim & 128 \\
Patch size & 1 \\
MLP ratio & 2.5 \\
$p_{drop}$ & 0.2 \\
Total parameters & 1.67M \\
Training steps & 200000 \\
Batch size & 64 \\
Optimizer & AdamW \\
Learning rate & $2 \times 10^{-4}$ \\
Weight decay & $10^{-2}$ \\
Max gradient norm & 1.0 \\
Floating point precision & float32 \\
\bottomrule
\end{tabular}
\caption{Neural network and training hyperparameters of the Diffusion Transformer architecture on the airfoil case.}
\label{tab:dit_airfoil_params}
\end{table}

\section{Selection of hyperparameters for the DiT-based flow-matching model on the 3D aircraft dataset}
\label{ap:hyper_aircraft}
Table \ref{tab:dit_aircraft_params} summarises the hyperparameters we used for the DiT architecture on the aircraft dataset. Like in previous flow-matching models, a cosine learning rate with no warmup has been used with a final learning rate value of $10^{-6}$.

\begin{table}[h]
\centering
\begin{tabular}{ll}
\toprule
\textbf{Parameter} & \textbf{Value} \\ 
\midrule
Num. blocks & 12 \\
Num. heads & 8 \\
Hidden dim & 256 \\
Patch size & 1 \\
MLP ratio & 4 \\
$p_{drop}$ & 0.2 \\
Total parameters & 81.3M \\
Training steps & 300000 \\
Batch size & 32 \\
Optimizer & AdamW \\
Learning rate & $2 \times 10^{-4}$ \\
Weight decay & $10^{-2}$ \\
Max gradient norm & 1.0 \\
Floating point precision & bfloat16 \\
\bottomrule
\end{tabular}
\caption{Neural network and training hyperparameters of the Diffusion Transformer architecture on the 3D aircraft case.}
\label{tab:dit_aircraft_params}
\end{table}

\begin{figure}[htb!]
\centering
\includegraphics[width=0.5\linewidth]{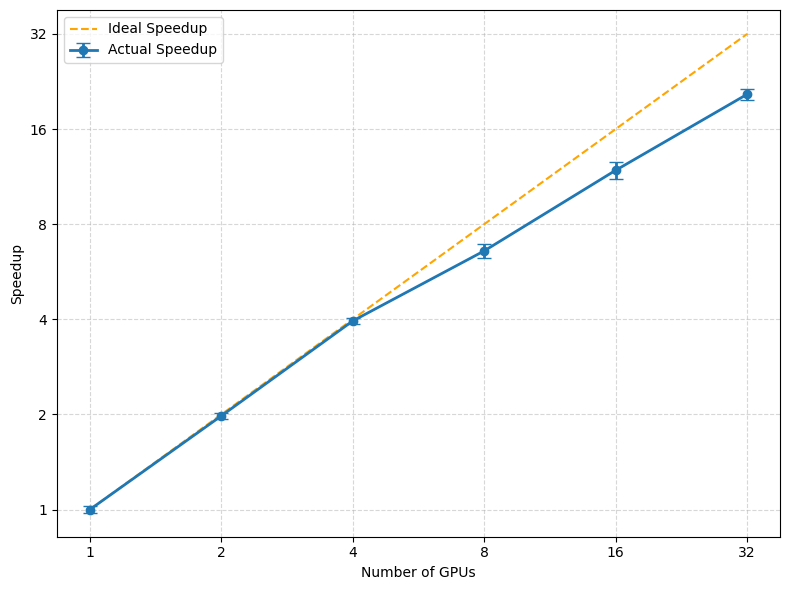}
\caption{Strong scaling speed-up of the DiT model for the aircraft dataset as a function of the number of GPUs.}
\label{fig:speedup}
\end{figure}

\section{Scaling training to many GPUs}
\label{ap:scalability}

To accelerate the training of the aircraft surrogate model, data-parallel training across multiple GPUs is employed. Specifically, an instance of the model is replicated on each GPU, and every training batch is partitioned into equally sized sub-batches that are distributed across devices. Each GPU computes the loss and corresponding gradients for its local sub-batch, after which gradients are synchronized and aggregated to update the model parameters. This parallelization strategy is implemented using the Hugging Face Accelerate library \cite{accelerate}.

\medskip\noindent 
To evaluate the scalability of this approach, a strong scaling study is conducted for the DiT model trained on the aircraft dataset, using up to 32 GPUs. In this configuration, the maximum number of GPUs is constrained by the global batch size, which is fixed to 32. Extending scalability beyond this limit would require combining the present data-parallel strategy with additional parallelization techniques, {such as model or pipeline parallelism.}
Figure~\ref{fig:speedup} reports the measured speed-up as a function of the number of GPUs, alongside the ideal linear scaling. Near-perfect scalability is observed up to 4 GPUs. However, for 8 GPUs and beyond, a clear deviation from ideal scaling emerges. This degradation is attributed to the underlying hardware topology of MN5: each compute node comprises 4 GPUs, such that configurations exceeding this number require inter-node communication, which introduces additional synchronization overhead and reduces overall efficiency.

\end{document}